\documentclass[11pt]{article}
\usepackage{acl2015}
\usepackage{times}
\usepackage[hyphens]{url}
\usepackage{hyperref}
\usepackage[hyphenbreaks]{breakurl}
\usepackage{latexsym}

\usepackage[T1, T2A]{fontenc}
\usepackage[utf8]{inputenc}

\usepackage{microtype}

\usepackage{enumitem}
\setlist[itemize]{noitemsep, nolistsep}
\setlist[enumerate]{noitemsep, nolistsep}
\usepackage[turkish,russian,english]{babel}

\usepackage{newunicodechar}

\newunicodechar{Ə}{\textazeriSchwa}
\newunicodechar{ə}{\textazerischwa}

\DeclareRobustCommand{\textazeriSchwa}{%
  {\fontencoding{T2A}\selectfont\symbol{"9A}}%
}
\DeclareRobustCommand{\textazerischwa}{%
  {\fontencoding{T2A}\selectfont\symbol{"BA}}%
}

\makeatletter
\expandafter\def\expandafter\@uclclist\expandafter
  {\@uclclist\textazerischwa\textazeriSchwa}
\makeatother

\usepackage{substitutefont}
\substitutefont{T2A}{\familydefault}{Tempora-TLF}
\title{Neural machine translation system for \\ Lezgian, Russian and Azerbaijani languages}

\author{Alidar Asvarov \thanks{\hspace{0.05cm} Native Lezgian speaker}\\
  Antiplagiat \\ Moscow, Russia \\
  Dagestan State University \\ Makhachkala, Russia \\
  {\tt alidar.asvarov@yandex.ru} \\\And
  Andrey Grabovoy \\
  Antiplagiat \\ Moscow, Russia \\
  Moscow Institue of Physics and Technology \\ Moscow, Russia \\
  {\tt grabovoy.av@phystech.edu} \\}

\date{}

\begin{document}
\maketitle
\begin{abstract}
  We release the first neural machine translation system for translation between Russian, Azerbaijani and the endangered Lezgian languages, as well as monolingual and parallel datasets collected and aligned for training and evaluating the system.
  Multiple experiments are conducted to identify how different sets of training language pairs and data domains can influence the resulting translation quality.
  We achieve BLEU scores of 26.14 for Lezgian-Azerbaijani, 22.89 for Azerbaijani-Lezgian, 29.48 for Lezgian-Russian and 24.25 for Russian-Lezgian pairs.
  The quality of zero-shot translation is assessed on a Large Language Model, showing its high level of fluency in Lezgian.
  However, the model often refuses to translate, justifying itself with its incompetence.
  % Claude~3.5 Large Language Model is also evaluated on zero-shot translation task and it shows high level of fluency in Lezgian, however it often refuses to translate, justifying itself with its incompetence.
  We contribute our translation model along with the collected parallel and monolingual corpora and sentence encoder for the Lezgian language.
\end{abstract}

\section{Introduction}

Northeast Caucasian or Nakh-Daghestanian language family contains 36 languages~\cite{glotolog}, however only few of them are covered by machine translation tools.
All of the languages in the family are considered low-resource, especially for limited amount of parallel corpora.
Google Translate in its recent update~\cite{new_in_google_translate} added more than 100 new languages and two of Northeast Caucasian languages are among them, namely Avar and Chechen.

In this work we focus on filling this gap for Lezgian language, which is spoken by Lezgians primarily in the Republic of Dagestan, Russia and in some provinces of the Republic of Azerbaijan.
Because Lezgians live in these two countries we select Russian and Azerbaijani languages as suitable targets for translation.
Lezgian belongs to Nakh-Daghestanian language family and is one of the official languages of the Republic of Dagestan. The language is based on Cyrillic script, however Latin-based alternative has been in use prior to 1938.

According to 2021 Russian census~\cite{ru_stat} and to 2019 Azerbaijani census~\cite{az_stat} approximately 454K people speak Lezgian, however this number is gradually declining and the language is classified as "vulnerable" by UNESCO's Atlas of the World's Languages in Danger~\cite{unesco_atlas}.
Conserving native languages of Russia is very important task that concerns Russian government, which opens state institutions to accomplish it~\cite{dom_narodov}.

As far as we know, no scientific research on neural machine translation (NMT) for Lezgian language have been published, even though there were attempts on building a NMT system between Lezgian and Russian~\footnote{\url{https://huggingface.co/leks-forever/nllb-200-distilled-600M}}.
In this work we research how different data sources and language pairs may influence the resulting NMT quality for Lezgian language and examine whether modern Large Language Models (LLMs) are already capable of producing satisfying zero-shot translations.
As a result of this work we make the following contribution:~\footnote{\url{https://github.com/Alidar40/lez-rus-azj-translator}}
\begin{itemize}
    \item A neural model for translation between Lezgian, Russian and Azerbaijani languages~\footnote{\url{https://huggingface.co/AlidarAsvarov/nllb-200-600M-lez-rus-azj}};
    \item A parallel Lezgian-Azerbaijani-Russian corpus and monolingual Lezgian corpus~\footnote{\url{https://huggingface.co/datasets/AlidarAsvarov/lezgi-rus-azer-corpus}};
    \item LaBSE-based sentence encoder for Lezgian.
\end{itemize}

% We also evaluate how a modern Large Language Model can perform on task of zero-shot translation for Lezgian language.

\section{Related work}
Оnly a very small number of the over 7000 languages of the
world are represented in the rapidly evolving language technologies and applications~\cite{joshi-etal-2020-state} and Lezgian belongs to this underrepresented family of languages.
The task of bridging the gap between low-resource and rich resource languages attracts a lot of efforts from researchers.
In recent years there were many successful attempts of building large-scale NMT systems for hundreds of languages, by utilizing large corpus mining and transfer learning~\cite{nllbteam2022language,bapna2022building}.
Especially when the data is limited it is interesting how different combinations of non-relative languages and data sources can contribute to transfer learning and impact translation quality for the low-resource language~\cite{Asvarov2024TheIO}.
Another promising approach is utilizing Large Language Models for machine translation task~\cite{zhu2024multilingualmachinetranslationlarge,xu2024paradigmshiftmachinetranslation}.
In recent years LLMs showed very powerful capabilities on all machine learning task, including zero-shot machine translation, despite having never
seen the intentionally-included translation examples~\cite{briakou2023searchingneedleshaystackrole}.

There are published small-scale monolingual and parallel corpora on Lezgian language. Namely, Lezgian wikipedia~\cite{wikidump} and OSCAR~\cite{oscar_2022arXiv220106642A}. 
There is a recently parsed Russian-Lezgian bible corpora~\footnote{\url{https://huggingface.co/datasets/leks-forever/bible-lezghian-russian}} and a model trained on it~\footnote{\url{https://huggingface.co/leks-forever/nllb-200-distilled-600M}}.

To the best of out knowledge, there are no works that release comparable size corpora, no works on developing translation system between Lezgian and Azerbaijani and no works on evaluating contemporary LLMs on the task of Lezgian language comprehension.

\section{Methodology}

\subsection{Data collection}

We use data from four sources:
\begin{itemize}
    \item \textit{Parallel sentences parsed from the Bible~\footnote{\url{https://www.bible.com/}}}. We choose "Oriental translation" (CARS) version for the Russian language, as we believe it is more coherent with the translation for the Lezgian language. Sentences have been aligned because some verses in one language are combined, while in another language that are two separate verses. In this case for the latter language verses are also combined.
    \item \textit{Parallel sentences parsed from the Quran~\footnote{\url{https://quranacademy.org/}}}. We choose  translation by Abu Adel for the Russian language as we believe it is more coherent with the translation for the Lezgian language.
    \item \textit{"Qusar, qusarilar"}~\cite{qusar_book} encyclopedia about Qusar region of Azerbaijan written in parallel Lezgian and Azerbaijani languages. As the straightforward programmatic text extraction from the distributed pdf-file is very challenging, the text was manually extracted, split by sentences and aligned using our sentence encoder (section  \ref{sec:sentence-encoder}). This data source is very important, as it is not limited by religious scope and contains texts from different domains such as history, politics, nature, culture, art, etc.
    \item Monolingual sentences parsed from "Lezgi Gazet"~\footnote{\url{https://lezgigazet.ru/}} -- the largest newspaper in Lezgian language. We download all available pdf-archives from the website, then extract the text with \texttt{pypdfium2}, normalize it with \texttt{sacremoses}, fix encoding if necessary, remove non-printable characters and split the text by sentences with \texttt{nltk}.
\end{itemize}

After parsing, clearing and aligninig we got total of 30069 multilingual sentence pairs (13617 for Bible, 6350 for Quran, and 10095 for Qusar encyclopedia) and almost 842K monolingual Lezgian sentences.

\subsection{Sentence Encoder}
\label{sec:sentence-encoder}

For text alignment purposes we create a sentence encoder from pre-trained LaBSE model~\cite{labse}.
First, we reduce the size of the original LaBSE model by removing all of the languages, but Russian and Azerbaijani~\footnote{Using the same method as for \url{https://huggingface.co/cointegrated/LaBSE-en-ru}}. Then we repeat the process, described in~\cite{dale2022neuralmachinetranslationerzya}.
Namely, we add new tokens for Lezgian language by training BPE~\cite{sennrich-etal-2016-neural} tokenizer over monolingual corpus.
The resulting model is fine-tuned with mono and parallel data (Bible and Quran subsets) for sentence pair classification and masked language modeling.

\subsection{Vocabulary expansion}

To leverage transfer learning we base our system on pre-trained \texttt{nllb-200-distilled-600} model trained on 200 languages, including Russian and Azerbaijani~\cite{nllbteam2022language}.
We extend its BPE vocabulary with extra Lezgian tokens using the method described in Section~\ref{sec:sentence-encoder} and add a new \texttt{lez\_Cyrl} language code.

\section{Experiments}

\subsection{Training Translation Models}
As we want to explore how different combinations of data  and different pairs of languages in the training set may influence the translation quality~\cite{Asvarov2024TheIO}, we start the first experiment by training the system to translate only between Lezgian and Azerbaijani languages, without Russian.

Then we hypothesize, that adding an additional language pair of Lezgian-Russian may improve the quality for Lezgian-Azerbaijani pair. Hence, for the second experiment we add an additional language pair of Lezgian-Russian with Bible and Quran sources.

In the third experiment we assume, that adding an additional pair of Azerbaijani-Russian to the training process may benefit the quality for directions with Lezgian Language. In this experiment we train the model with the same data as in the second experiment, but also add translation between Azerbaijani and Russian to the training objective.

In the fourth experiment we utilize pre-trained \texttt{nllb-200-distilled-1.3B} model to translate the Qusar data source from Azerbaijani to Russian.
This form of offline back-translation allows us to use this new data to train additional Lezgian-Russian language pair on the domain of the Qusar encyclopedia.
Even though this method doesn't generate perfect translations, it is shown to be useful~\cite{sennrich-etal-2016-improving,poncelas-etal-2018-investigating} and gives a significant increase in translation quality, especially when dealing with low-resource languages, such as Lezgian.
In this experiment we train our final NMT system on all of the parallel data and on all language pair combinations.

\subsection{Test Data}

We evaluate our models with a holdout dataset of size 1000, which was derived from splitting the corpus in a stratified fashion.
As the are no original Russian translation for the Qusar Encyclopedia we utilize Google Translator~\footnote{\url{https://translate.google.com/}} to translate from Azerbaijani language and generate test sentences in Russian for this source.
This is not a perfect solution for obtaining evaluation data, however this allows us to estimate the quality of models.

\subsection{Automatic metrics}

In this work we report BLEU~\cite{papineni-etal-2002-bleu} and ChrF++ scores~\cite{popovic-2017-chrf-plus-plus}.
Due to its nature, ChrF++ may be more plausible, as all of the languages in this work are morphologicaly rich.
BLEU scores for all of the experiments are given in Tables \ref{tab:bleu_1exp}, \ref{tab:bleu_2exp}, \ref{tab:bleu_3exp} and \ref{tab:bleu_4exp}.
ChrF++ scores are given in Appendix~\ref{app:chrf_scores}.

First, we see that аll our hypotheses were not confirmed.
Adding new language pairs does not help to improve quality of NLLB fine-tuning, despite identical data sources.
% For fine-tuning NLLB model adding new language pairs does not help to improve quality, despite the same data sources.

In the second experiment translation quality between Lezgian and Azerbaijani is relatively similar in both directions.
In Russian-Lezgian pair we see a clear distinction between BLEU and ChrF++ scores: the first one shows that on religious texts \texttt{ru-lez} direction is more challenging than \texttt{lez-ru}, but the latter shows that there is no difference.
However, both metrics agree that \texttt{lez-ru} direction is more challenging on unseen domain of Qusar. This could be explained by much richer vocabulary of Russian language for this data source.
We observe the same pattern for the third experiment.
In the last experiment there is a big gap in scores for translation between Russian and Azerbaijani languages on Qusar dataset.
This could be explained by the fact that both of this languages have already been represented in the base NLLB model.

\begin{table}
\small
\centering
\begin{tabular}{l|rrrr}
\hline
\textbf{} & \textbf{All} & \textbf{Bible} & \textbf{Quran} & \textbf{Qusar} \\
\hline
lez-az & 24.69 & 21.93 & 30.12 & 25.18 \\
az-lez & 23.08 & 19.86 & 27.38 & 25.14 \\
\hline
\end{tabular}
\caption{\label{tab:bleu_1exp} BLEU scores for the first experiment: training on just Lezgian and Azerbaijani sources.}
\end{table}

\begin{table}
\small
\centering
\begin{tabular}{l|rrrr}
\hline
\textbf{} & \textbf{All} & \textbf{Bible} & \textbf{Quran} & \textbf{Qusar} \\
\hline
lez-az & 25.58 & 23.01 & 31.55 & 25.28 \\
az-lez & 23.64 & 20.26 & 28.04 & 26.02 \\
lez-ru & 26.38 & 27.65 & 41.80 & 6.51  \\
ru-lez & 22.47 & 21.98 & 31.45 & 12.61 \\
\hline
\end{tabular}
\caption{\label{tab:bleu_2exp} BLEU scores for the second experiment: training Lezgian-Azerbaijani pair with addition of Russian-Lezgian pair on religious data.}
\end{table}

\begin{table}[!ht]
\small
\centering
\begin{tabular}{l|rrrr}
\hline
\textbf{} & \textbf{All} & \textbf{Bible} & \textbf{Quran} & \textbf{Qusar} \\
\hline
lez-az & 25.76 & 23.28 & 31.01 & 25.93 \\
az-lez & 22.99 & 20.06 & 26.92 & 25.00 \\
lez-ru & 26.11 & 27.68 & 40.52 & 6.76  \\
ru-lez & 22.95 & 22.81 & 31.60 & 12.58 \\
ru-az  & 27.38 & 26.70 & 32.81 & 24.09 \\
az-ru  & 27.64 & 30.16 & 37.29 & 11.33 \\
\hline
\end{tabular}
\caption{\label{tab:bleu_3exp} BLEU scores for the third experiment: training with an additional Russian-Azerbaijani direction.}
\end{table}

\begin{table}
\small
\centering
\begin{tabular}{l|rrrr}
\hline
\textbf{} & \textbf{All} & \textbf{Bible} & \textbf{Quran} & \textbf{Qusar} \\
\hline
lez-az & 26.14 & 24.21 & 30.77 & 25.85 \\
az-lez & 22.89 & 20.27 & 27.29 & 23.66 \\
lez-ru & 29.48 & 27.61 & 41.42 & 21.35 \\
ru-lez & 24.25 & 22.10 & 31.78 & 20.31 \\
ru-az  & 31.65 & 25.73 & 32.93 & 41.10 \\
az-ru  & 33.63 & 28.78 & 36.83 & 40.46 \\
\hline
\end{tabular}
\caption{\label{tab:bleu_4exp} BLEU scores for the fourth experiment: offline back-translation of Qusar data source to Russian and training on whole corpus and all directions.}
\end{table}

\subsection{Comparison with Claude~3.5 Sonnet Large Language Model}

In this research we try to utilize Claude~3.5 Sonnet~\cite{claude_sonnet} to generate zero-shot translations on the test set.
We choose this model as it is currently one of the top performing models with free (but limited) access and on our first approach it did not refuse to work with Lezgian language.

We select first 100 sentences from the Qusar source of the test set.
This data source is chosen as it is more challenging and our models perform on it worse, than on religious texts.
We've trimmed the set to 100 sentences because of Claude's limit for maximum messages count and length.
We form a csv-file for each source language, then upload it to the Claude's interface and ask it to translate with the following prompt: \texttt{"This is a csv file with sentences in Lezgian language. Please translate all of them in Russian language"}. We change languages in the prompt depending on the source and target translation. Then we download and parse Claude's answers and calculate BLEU and ChrF++ scores, which are shown in Table \ref{tab:bleu_claude}.

However, Claude refused to perform translation to Lezgian language, justifying itself with its incompetence with the language.
This behaviour is very strange, considering that Claude has agreed to translate couple of standalone sentences and performed the translation on a good level.
We tried many different prompts and revisited the system for many days, but all of the attempts were futile.

We can conclude that Claude has reached fairly high level of Lezgian language acquisition, but its use is limited by Anthropic's safety guidelines.

We show couple of examples of translated sentences in Table~\ref{tab:main_examples}.
Let us recall, that the sentences in Russian were obtained using Google Translate.
In the first example our translator and Google make the same mistake of translating \texttt{az: divan tutur/lez: дуван куьн} [\texttt{eng: to judge/to punish}, rus: \texttt{учинять расправу}] as \texttt{lez: есирда куьн/rus: попасть в плен} [eng: \texttt{to be captured}].
Our model changed a little bit the meaning of the sentence for \texttt{lez-az} direction.
Claude has translated perfectly for all directions, except for \texttt{az-lez}, but still the meaning is preserved.
In the second example we see that Google omitted the information of Hasan's origin, while Claude did not.

\begin{table}
\small
\centering
\begin{tabular}{l|rrr}
\hline
\textbf{Direction} & \textbf{BLEU} & \textbf{ChrF++} \\
\hline
lez-az & 34.66 & 53.8 \\
az-lez & Refused & Refused \\
lez-ru & 21.20 & 44.37 \\
ru-lez & Refused & Refused \\ 
\hline
\end{tabular}
\caption{\label{tab:bleu_claude} BLEU and ChrF++ scores for Claude~3.5 Sonnet.}
\end{table}

Examples of translations are given in Table \ref{tab:main_examples} and in Appendix \ref{app:examples}.

\begin{table*}[ht!]
\small
\centering
\begin{tabular}{|l|p{0.7\linewidth}|}
\hline
source (az)          & 1741-ci ilin payızında Nadir şah Müşkür ləzgilərinə hücum edib onlara divan tutur.  \\
source (ru)         & Осенью 1741 года Надир-шах напал на мушкурских лезгин и взял их в плен.  \\
source (lez)         & 1741-йисан зулуз Надир шагьди Муьшкуьрдин лезгийрал вегьена, абуруз дуван кьунай.  \\
translation (az-lez) & 1741-йисан зулуз Надир шагь Муьшкуьрди лезгийриз басрух гана ва абур есирда кьуна. \\
translation (lez-az) & 1741-ci ilin payızında Nadir şah Müşkür ləzgilərinin üstünə hücum çəkib onlara qarşı qəhrəmanlıqla döyüşmüşdür. \\
translation (ru-lez) & 1741-йисан зулуз Надир шагьди Муьшкуьр лезгийриз басрух гана, абур есирда кьунай. \\
translation (lez-ru) & Осенью 1741 года Надир-шах напал на лезгин Мушкура и обесчестил их. \\
translation (Claude) (az-lez) & 1741 йисан зулуз Надир шагьди Муьшкуьрдин лезгийрал гьужум авуна абур кукIварна. \\
translation (Claude) (lez-az) & 1741-ci ilin payızında Nadir şah Müşkür ləzgilərinə hücum edib onlara divan tutdu. \\
translation (Claude) (ru-lez) & 1741 йисан зулуз Надир-шагьди мушкур лезгийрал гьужумна ва абур есирвиле кьуна. \\
translation (Claude) (lez-ru) & Осенью 1741 года Надир-шах напал на лезгин Мюшкюра и учинил над ними расправу. \\
\hline
source (az)          & Rus yazıçısı Vladimir Dal ədalət uğrunda mübarizə aparmış gədəzeyxürlü qaçaq Həsən haqqında "Ləzgi Həsən" adlı povest yazmışdır.  \\
source (ru)         & Русский писатель Владимир Даль написал рассказ "Лезги Хасан" о Хасане, беглеце, боровшемся за справедливость.  \\
source (lez)         & Урусрин машгьур кхьираг Владимир Дала гьахъ патал женг чӀугур гадацӀийихуьруьнви къачагъ Гьасанакай "Лезги Гьасан" тӀвар алай повесть кхьенай.  \\
translation (az-lez) & Владимир Дала гьахъсузвиляй женг чӀугвазвай гадацӀийихуьруьнви къачагъ Гьасанакай "Лезги Гьасан" тӀвар ганвай повесть кхьенай. \\
translation (lez-az) & Məşhur rus yazıçısı Vladimir Dal haqqı qorumaq üçün mübarizə aparan Gədəzeyxürlü qaçaq Həsən haqqında "Ləzgi Həsən" povestini yazmışdır. \\
translation (ru-lez) & Россиядин кхьираг Владимир Дала адалатлувал патал женг чӀугвазвай къачагъ Гьасанакай "Лезги Гьасан" тӀвар ганвай повесть кхьенай. \\
translation (lez-ru) & Известный российский писатель Владимир Даль написал рассказ "Лезгинский Гасан" о Гадазейхурском похитителе Хасане, боровшемся за свободу. \\
translation (Claude) (az-lez) & Урус кхьираг Владимир Дала дуьзвал патал женг чIугур КьеӀнихуьруьн къачагъ Гьасанакай "Лезги Гьасан" тIвар алай повесть кхьенай. \\
translation (Claude) (lez-az) & Rus yazıçısı Vladimir Dal haqq uğrunda mübarizə aparmış qadatsiyixürlü qaçaq Həsən haqqında "Ləzgi Həsən" adlı povest yazmışdı. \\
translation (Claude) (ru-lez) & Урус кхьираг Владимир Дала 'Лезги Гьасан' тIвар алай гьикая кхьена. Ам дуьзвал патал женг чIугур катнавай Гьасанакай я. \\
translation (Claude) (lez-ru) & Известный русский писатель Владимир Даль написал повесть под названием 'Лезгин Гасан' о разбойнике из Гаджи-Нового Села, боровшемся за справедливость \\
\hline
\end{tabular}
\caption{\label{tab:main_examples} Examples of translations for Qusar data source.}
\end{table*}

\section{Conclusion}
\label{sec:conclusion}

We present a neural machine translation system which can translate between endangered Lezgian, Russian and Azerbaijani languages.

For this work we have collected 30K parallel sentences for Lezgian-Azerbaijani language pair, 20K of these sentences have Russian version, and a corpus of almost 842K monolingual Lezgian sentences and a sentence encoder.

We finetune pre-trained NLLB model on our data and achieve BLEU scores of 26.14 for Lezgian-Azerbaijani, 22.89 for Azerbaijani-Lezgian, 29.48 for Lezgian-Russian and 24.25 for Russian-Lezgian pairs.

Our research shows, that adding additional language pairs doesn't improve the quality, despite being trained on the same data source.

We evaluate the quality of Claude~3.5 model for zero-shot translation and conclude that it has achieved high level of Lezgian language aquisition.
But Anthropic's safety guidelines often force the model to refuse to perform the translation, justifing it by its incompetence.

\section{Future Work}
\label{sec:future_work}

In future work, we plan to focus on creating a larger, more diverse and better quality parallel corpora.
Employing contemporary LLMs, especially for low-resource languages, is a very promising approach, which we want to research in more detail.

\bibliography{main}

\begin{thebibliography}{}

\bibitem[\protect\citename{{Abadji} \bgroup et al.\egroup
  }2022]{oscar_2022arXiv220106642A}
Julien {Abadji}, Pedro {Ortiz Suarez}, Laurent {Romary}, and Beno{\^\i}t
  {Sagot}.
\newblock 2022.
\newblock {Towards a Cleaner Document-Oriented Multilingual Crawled Corpus}.
\newblock {\em arXiv e-prints}, page arXiv:2201.06642, January.

\bibitem[\protect\citename{Anthropic}2024]{claude_sonnet}
Anthropic.
\newblock 2024.
\newblock Claude 3.5 sonnet.
\newblock \url{https://www.anthropic.com/news/claude-3-5-sonnet}, June.
\newblock Accessed: 01 October 2024.

\bibitem[\protect\citename{Asvarov and Grabovoy}2024]{Asvarov2024TheIO}
Alidar Asvarov and Andrey Grabovoy.
\newblock 2024.
\newblock The impact of multilinguality and tokenization on statistical machine
  translation.
\newblock {\em 2024 35th Conference of Open Innovations Association (FRUCT)},
  pages 149--157.

\bibitem[\protect\citename{Azstat}2019]{az_stat}
Azstat.
\newblock 2019.
\newblock Ahalinin milli tarkibi (ahalinin siyahiyaalinmalarinin malumatlarina
  asasan). azarbaycan respublikasi dovlat statistika komitasi [national
  composition of the population (based on data from population censuses). state
  statistics committee of the republic of azerbaijan].
\newblock Accessed: 01 October 2024.

\bibitem[\protect\citename{Bapna \bgroup et al.\egroup
  }2022]{bapna2022building}
Ankur Bapna, Isaac Caswell, Julia Kreutzer, Orhan Firat, Daan van Esch, Aditya
  Siddhant, Mengmeng Niu, Pallavi Baljekar, Xavier Garcia, Wolfgang Macherey,
  et~al.
\newblock 2022.
\newblock Building machine translation systems for the next thousand languages.
\newblock {\em arXiv preprint arXiv:2205.03983}.

\bibitem[\protect\citename{Briakou \bgroup et al.\egroup
  }2023]{briakou2023searchingneedleshaystackrole}
Eleftheria Briakou, Colin Cherry, and George Foster.
\newblock 2023.
\newblock Searching for needles in a haystack: On the role of incidental
  bilingualism in palm's translation capability.

\bibitem[\protect\citename{Dale}2022]{dale2022neuralmachinetranslationerzya}
David Dale.
\newblock 2022.
\newblock The first neural machine translation system for the erzya language.

\bibitem[\protect\citename{Feng \bgroup et al.\egroup }2022]{labse}
Fangxiaoyu Feng, Yinfei Yang, Daniel Cer, Naveen Arivazhagan, and Wei Wang.
\newblock 2022.
\newblock Language-agnostic {BERT} sentence embedding.
\newblock In {\em Proceedings of the 60th Annual Meeting of the Association for
  Computational Linguistics (Volume 1: Long Papers)}, pages 878--891, Dublin,
  Ireland, May. Association for Computational Linguistics.

\bibitem[\protect\citename{Glottolog}2024]{glotolog}
Glottolog.
\newblock 2024.
\newblock Family: Nakh-daghestanian.
\newblock \url{https://glottolog.org/resource/languoid/id/nakh1245}.
\newblock Accessed: 01 October 2024.

\bibitem[\protect\citename{Google}2024]{new_in_google_translate}
Google.
\newblock 2024.
\newblock What’s new in google translate: More than 100 new languages.
\newblock \url{https://support.google.com/translate/answer/15139004?hl=en}.
\newblock Accessed: 01 October 2024.

\bibitem[\protect\citename{Joshi \bgroup et al.\egroup
  }2020]{joshi-etal-2020-state}
Pratik Joshi, Sebastin Santy, Amar Budhiraja, Kalika Bali, and Monojit
  Choudhury.
\newblock 2020.
\newblock The state and fate of linguistic diversity and inclusion in the {NLP}
  world.
\newblock In Dan Jurafsky, Joyce Chai, Natalie Schluter, and Joel Tetreault,
  editors, {\em Proceedings of the 58th Annual Meeting of the Association for
  Computational Linguistics}, pages 6282--6293, Online, July. Association for
  Computational Linguistics.

\bibitem[\protect\citename{Kerimova}2011]{qusar_book}
Sedaget Kerimova.
\newblock 2011.
\newblock {\em Qusar, qusarilar [Gusar, gusaris]}.
\newblock Ziya, Baku.

\bibitem[\protect\citename{NLLB~Team \bgroup et al.\egroup
  }2022]{nllbteam2022language}
Marta R. Costa-jussà NLLB~Team, James Cross, Onur Çelebi, Maha Elbayad,
  Kenneth Heafield, Kevin Heffernan, Elahe Kalbassi, Janice Lam, Daniel Licht,
  Jean Maillard, Anna Sun, Skyler Wang, Guillaume Wenzek, Al~Youngblood, Bapi
  Akula, Loic Barrault, Gabriel~Mejia Gonzalez, Prangthip Hansanti, John
  Hoffman, Semarley Jarrett, Kaushik~Ram Sadagopan, Dirk Rowe, Shannon Spruit,
  Chau Tran, Pierre Andrews, Necip~Fazil Ayan, Shruti Bhosale, Sergey Edunov,
  Angela Fan, Cynthia Gao, Vedanuj Goswami, Francisco Guzmán, Philipp Koehn,
  Alexandre Mourachko, Christophe Ropers, Safiyyah Saleem, Holger Schwenk, and
  Jeff Wang.
\newblock 2022.
\newblock No language left behind: Scaling human-centered machine translation.

\bibitem[\protect\citename{Papineni \bgroup et al.\egroup
  }2002]{papineni-etal-2002-bleu}
Kishore Papineni, Salim Roukos, Todd Ward, and Wei-Jing Zhu.
\newblock 2002.
\newblock {B}leu: a method for automatic evaluation of machine translation.
\newblock In Pierre Isabelle, Eugene Charniak, and Dekang Lin, editors, {\em
  Proceedings of the 40th Annual Meeting of the Association for Computational
  Linguistics}, pages 311--318, Philadelphia, Pennsylvania, USA, July.
  Association for Computational Linguistics.

\bibitem[\protect\citename{Poncelas \bgroup et al.\egroup
  }2018]{poncelas-etal-2018-investigating}
Alberto Poncelas, Dimitar Shterionov, Andy Way, Gideon Maillette~de
  Buy~Wenniger, and Peyman Passban.
\newblock 2018.
\newblock Investigating backtranslation in neural machine translation.
\newblock In Juan~Antonio P{\'e}rez-Ortiz, Felipe S{\'a}nchez-Mart{\'\i}nez,
  Miquel Espl{\`a}-Gomis, Maja Popovi{\'c}, Celia Rico, Andr{\'e} Martins,
  Joachim Van~den Bogaert, and Mikel~L. Forcada, editors, {\em Proceedings of
  the 21st Annual Conference of the European Association for Machine
  Translation}, pages 269--278, Alicante, Spain, May.

\bibitem[\protect\citename{Popovi{\'c}}2017]{popovic-2017-chrf-plus-plus}
Maja Popovi{\'c}.
\newblock 2017.
\newblock chr{F}++: words helping character n-grams.
\newblock In Ond{\v{r}}ej Bojar, Christian Buck, Rajen Chatterjee, Christian
  Federmann, Yvette Graham, Barry Haddow, Matthias Huck, Antonio~Jimeno Yepes,
  Philipp Koehn, and Julia Kreutzer, editors, {\em Proceedings of the Second
  Conference on Machine Translation}, pages 612--618, Copenhagen, Denmark,
  September. Association for Computational Linguistics.

\bibitem[\protect\citename{Rosstat}2021]{ru_stat}
Rosstat.
\newblock 2021.
\newblock Vladenie yazykami i ispol'zovanie yazykov naseleniem rossijskoj
  federacii soglasno perepisi naseleniya 2021 goda [language proficiency and
  use of languages by the population of the russian federation according to the
  2021 population census].
\newblock
  \url{https://web.archive.org/web/20230326024701/https://rosstat.gov.ru/storage/mediabank/Tom5_tab4_VPN-2020.xlsx}.
\newblock Accessed: 01 October 2024.

\bibitem[\protect\citename{Russian{\ }Government}2019]{dom_narodov}
Russian{\ }Government.
\newblock 2019.
\newblock Rasporyazhenie pravitel'stva rf o sozdanii fgbu \"dom narodov
  rossii\" [order of the government of the russian federation on the
  establishment of the federal state budgetary institution "house of the
  peoples of russia"].
\newblock Accessed: 01 October 2024.

\bibitem[\protect\citename{Sennrich \bgroup et al.\egroup
  }2016a]{sennrich-etal-2016-improving}
Rico Sennrich, Barry Haddow, and Alexandra Birch.
\newblock 2016a.
\newblock Improving neural machine translation models with monolingual data.
\newblock In Katrin Erk and Noah~A. Smith, editors, {\em Proceedings of the
  54th Annual Meeting of the Association for Computational Linguistics (Volume
  1: Long Papers)}, pages 86--96, Berlin, Germany, August. Association for
  Computational Linguistics.

\bibitem[\protect\citename{Sennrich \bgroup et al.\egroup
  }2016b]{sennrich-etal-2016-neural}
Rico Sennrich, Barry Haddow, and Alexandra Birch.
\newblock 2016b.
\newblock Neural machine translation of rare words with subword units.
\newblock In Katrin Erk and Noah~A. Smith, editors, {\em Proceedings of the
  54th Annual Meeting of the Association for Computational Linguistics (Volume
  1: Long Papers)}, pages 1715--1725, Berlin, Germany, August. Association for
  Computational Linguistics.

\bibitem[\protect\citename{UNESCO}2010]{unesco_atlas}
UNESCO.
\newblock 2010.
\newblock Unesco atlas of the world's languages in danger (pdf).
\newblock Accessed: 01 October 2024.

\bibitem[\protect\citename{Wikimedia}2024]{wikidump}
Wikimedia.
\newblock 2024.
\newblock Wikimedia downloads.
\newblock Accessed: 01 October 2024.

\bibitem[\protect\citename{Xu \bgroup et al.\egroup
  }2024]{xu2024paradigmshiftmachinetranslation}
Haoran Xu, Young~Jin Kim, Amr Sharaf, and Hany~Hassan Awadalla.
\newblock 2024.
\newblock A paradigm shift in machine translation: Boosting translation
  performance of large language models.

\bibitem[\protect\citename{Zhu \bgroup et al.\egroup
  }2024]{zhu2024multilingualmachinetranslationlarge}
Wenhao Zhu, Hongyi Liu, Qingxiu Dong, Jingjing Xu, Shujian Huang, Lingpeng
  Kong, Jiajun Chen, and Lei Li.
\newblock 2024.
\newblock Multilingual machine translation with large language models:
  Empirical results and analysis.

\end{thebibliography}
\bibliographystyle{acl}

\appendix
\onecolumn

\section{ChrF++ scores}
\label{app:chrf_scores}

\begin{table*}[ht!]
\small
\centering
\begin{tabular}{l|rrrr}
\hline
\textbf{} & \textbf{All} & \textbf{Bible} & \textbf{Quran} & \textbf{Qusar} \\
\hline
lez-az & 47.89 & 46.10 & 53.63 & 46.55 \\
az-lez & 49.27 & 46.68 & 54.03 & 49.80 \\
\hline
\end{tabular}
\caption{\label{tab:bleu} ChrF++ scores for the first experiment: training on just Lezgian and Azerbaijani sources.}
\end{table*}

\begin{table*}[ht!]
\small
\centering
\begin{tabular}{l|rrrr}
\hline
\textbf{} & \textbf{All} & \textbf{Bible} & \textbf{Quran} & \textbf{Qusar} \\
\hline
lez-az & 48.21 & 46.68 & 54.08 & 46.34 \\
az-lez & 49.06 & 46.22 & 54.16 & 49.77 \\
lez-ru & 45.12 & 48.50 & 59.98 & 27.17 \\
ru-lez & 47.88 & 47.99 & 57.57 & 38.45 \\
\hline
\end{tabular}
\caption{\label{tab:bleu} ChrF++ scores for the second experiment: training Lezgian-Azerbaijani pair with addition of Russian-Lezgian pair on religious data.}
\end{table*}

\begin{table*}[ht!]
\small
\centering
\begin{tabular}{l|rrrr}
\hline
\textbf{} & \textbf{All} & \textbf{Bible} & \textbf{Quran} & \textbf{Qusar} \\
\hline
lez-az & 48.63 & 47.01 & 53.99 & 47.32 \\
az-lez & 48.85 & 45.92 & 53.77 & 49.84 \\
lez-ru & 44.36 & 47.83 & 58.93 & 26.77 \\
ru-lez & 61.80 & 55.11 & 63.90 & 72.90 \\
ru-az  & 51.28 & 50.45 & 56.08 & 48.99 \\
az-ru  & 47.54 & 50.24 & 56.92 & 35.26 \\
\hline
\end{tabular}
\caption{\label{tab:bleu} ChrF++ scores for the third experiment: training with an additional Russian-Azerbaijani direction.}
\end{table*}

\begin{table*}[ht!]
\small
\centering
\begin{tabular}{l|rrrr}
\hline
\textbf{} & \textbf{All} & \textbf{Bible} & \textbf{Quran} & \textbf{Qusar} \\
\hline
lez-az & 48.62 & 47.17 & 53.95 & 47.02 \\
az-lez & 48.37 & 46.19 & 53.65 & 47.61 \\
lez-ru & 48.74 & 47.76 & 59.33 & 41.85 \\
ru-lez & 49.55 & 47.95 & 57.49 & 45.16 \\
ru-az  & 54.97 & 49.46 & 55.50 & 63.62 \\
az-ru  & 54.71 & 49.21 & 56.32 & 62.54 \\
\hline
\end{tabular}
\caption{\label{tab:bleu} ChrF++ scores for the fourth experiment: offline back-translation of Qusar data source to Russian and training on whole corpus and all directions.}
\end{table*}

% \section{Model Hyperparameters}
% \label{app:hyperparams}

\clearpage
\pagebreak
\section{Translation Examples}
\label{app:examples}

\begin{table*}[ht!]
\small
\centering
\begin{tabular}{|l|p{0.7\linewidth}|}
\hline
source (az)          & Elə həmin günlərdə Yerusəlimdən Antakyaya bəzi peyğəmbərlər gəldi.  \\
source (ru)         & В это время из Иерусалима в Антиохию пришли несколько пророков. \\
source (lez)         & И чӀавуз Ярусалимдай АнтӀиахдиз пайгъамбарар атана.  \\
translation (az-lez) & Гьа йикъара АнтӀиахдиз Ярусалимдай пайгъамбарар атана. \\
translation (lez-az) & Bu günlər Yerusəlimdən Antakyaya peyğəmbərlər gəldi. \\
translation (ru-lez) & И йикъара АнтӀиахдиз Ярусалимдай пайгъамбарар атана. \\
translation (lez-ru) & В то время из Иерусалима в Антиохию пришли пророки. \\
\hline
source (az)          & İndi siz onunla gördüyü şey barəsində mübahisəmi edirsiniz?  \\
source (ru)         & Неужели вы станете спорить с ним то, что он видел? \\
source (lez)         & Бес, куьне гьуьжетзавани адахъ адаз аквазвайдан гьакъиндай?!  \\
translation (az-lez) & Бес, куьне ада кьунвай кардин патахъай гьуьжетарзавани?! \\
translation (lez-az) & Onun gördüyü şeylər barədə mübahisəmi edirsiniz? \\
translation (ru-lez) & Бес, куьне ада кьунвай кардин патахъай гьуьжетзавани - \\
translation (lez-ru) & Неужели вы препираетесь с ним относительно того, что он видит? \\
\hline
source (az)          & Bu məlumat dilçilik elmində mühüm hadisəyə çevrilmişdir. \\
source (ru)         & Эта информация стала важным событием в лингвистике. \\
source (lez)         & И малумат чӀалан илимда чӀехи вакъиадиз элкъвенай.  \\
translation (az-lez) & И малуматар лингвистикада важиблу вакъиадиз элкъвена. \\
translation (lez-az) & Bu məlumat linqvistika elminə böyük töhfə gətirmişdir. \\
translation (ru-lez) & И малумат лингвистикада чӀехи вакъиадиз элкъвена. \\
translation (lez-ru) & Это стало большим событием в языкознании. \\
\hline
source (az)          & Bu təşkilat evdar qadınları işə cəlb etmək üçün tikiş arteli təşkil etmiş, savadsızlığı ləğv etmə kursları yaratmışdı. \\
source (ru)         & Эта организация организовала швейную артель для привлечения к работе домохозяек, создала курсы по ликвидации неграмотности. \\
source (lez)         & И тешкилатди кайванияр патал пекер цудай артел тешкилнай, савадсузвал тергдай курсар ачухарнай.  \\
translation (az-lez) & КӀвалин дишегьлияр кӀвалахдив агудун патал ада хранвай пекер хрунин тарсар гузвай. \\
translation (lez-az) & Həmin təşkilat toyxanaçılar üçün geyim toxuyan artellə işləmiş, savadsızlığı ləğvetmə kurslarını yaratmışdı. \\
translation (ru-lez) & Адан регьбервилик кваз хуьруьн майишатдин кӀвалахар теблигъ авун патал хранвай артел тешкилна савадсузвал терг авун патал курсар арадиз гъана. \\
translation (lez-ru) & Организация организовала артели для вышивки одежды и курсы ликвидации неграмотности для домохозяйств. \\
\hline
\end{tabular}
\caption{\label{tab:main_examples} Examples of translations from the test set}
\end{table*}

\end{document}